\documentclass[10pt,letterpaper]{article}

\usepackage{cogsci}
\usepackage{pslatex}
\usepackage{apacite}
\usepackage{graphicx}
\usepackage{subcaption}
\usepackage{natbib}
\usepackage[super]{nth}
\usepackage[hidelinks]{hyperref}
\usepackage{amsmath}

\usepackage[font=small]{caption}
\renewcommand\bibliographytypesize{\scriptsize}

\title{Learning Inductive Biases with Simple Neural Networks}

\author{{\large \bf Reuben Feinman (reuben.feinman@nyu.edu)} \\
  Center for Neural Science \\
  New York University
  \And {\large \bf Brenden M. Lake (brenden@nyu.edu)} \\
  Department of Psychology and Center for Data Science \\
  New York University}

\begin{document}
\maketitle

\begin{abstract}
    People use rich prior knowledge about the world in order to efficiently learn
new concepts. These priors--also known as ``inductive biases"--pertain to the
space of internal models considered by a learner, and they help the learner
make inferences that go beyond the observed data. A recent study found that
deep neural networks optimized for object recognition develop the shape bias
\citep{Ritter2017}, an inductive bias possessed by children that plays an
important role in early word learning. However, these networks use unrealistically
large quantities of training data, and the conditions required for these biases to develop are not
well understood. Moreover, it is unclear how the learning dynamics of these
networks relate to developmental processes in childhood. We
investigate the development and influence of the shape bias in neural networks
using controlled datasets of abstract patterns and synthetic images, allowing
us to systematically vary the quantity and form of the experience provided to
the learning algorithms. We find that simple neural networks develop a shape
bias after seeing as few as 3 examples of 4 object categories. The development of these biases predicts the onset of vocabulary acceleration in our networks, consistent with
the developmental process in children.

\end{abstract}

Humans possess the remarkable ability to learn a new concept from seeing just a
few examples. A child can learn the meaning of a new word such as ``fork" after observing only one or a handful of different forks \citep{Bloom2000}. In contrast, state-of-the-art artificial learning systems use
hundreds or thousands of examples per class when learning to recognize the same objects
\citep[e.g.,][]{Krizhevsky2012, Szegedy2015}. Consequently, significant
effort is ongoing to understand what cognitive and neural mechanisms enable
efficient concept learning \citep{Lake2017}. In this paper, we perform a series
of developmentally-informed neural network experiments to study the
computational basis of efficient word learning.\footnote{All experiments can be
reproduced using the code repository located at
\url{http://github.com/rfeinman/learning-to-learn}.}

\begin{figure}[h!]
    \begin{center}
    	\begin{subfigure}[b]{0.23\textwidth}
            \begin{center}
                \includegraphics[width=\linewidth]{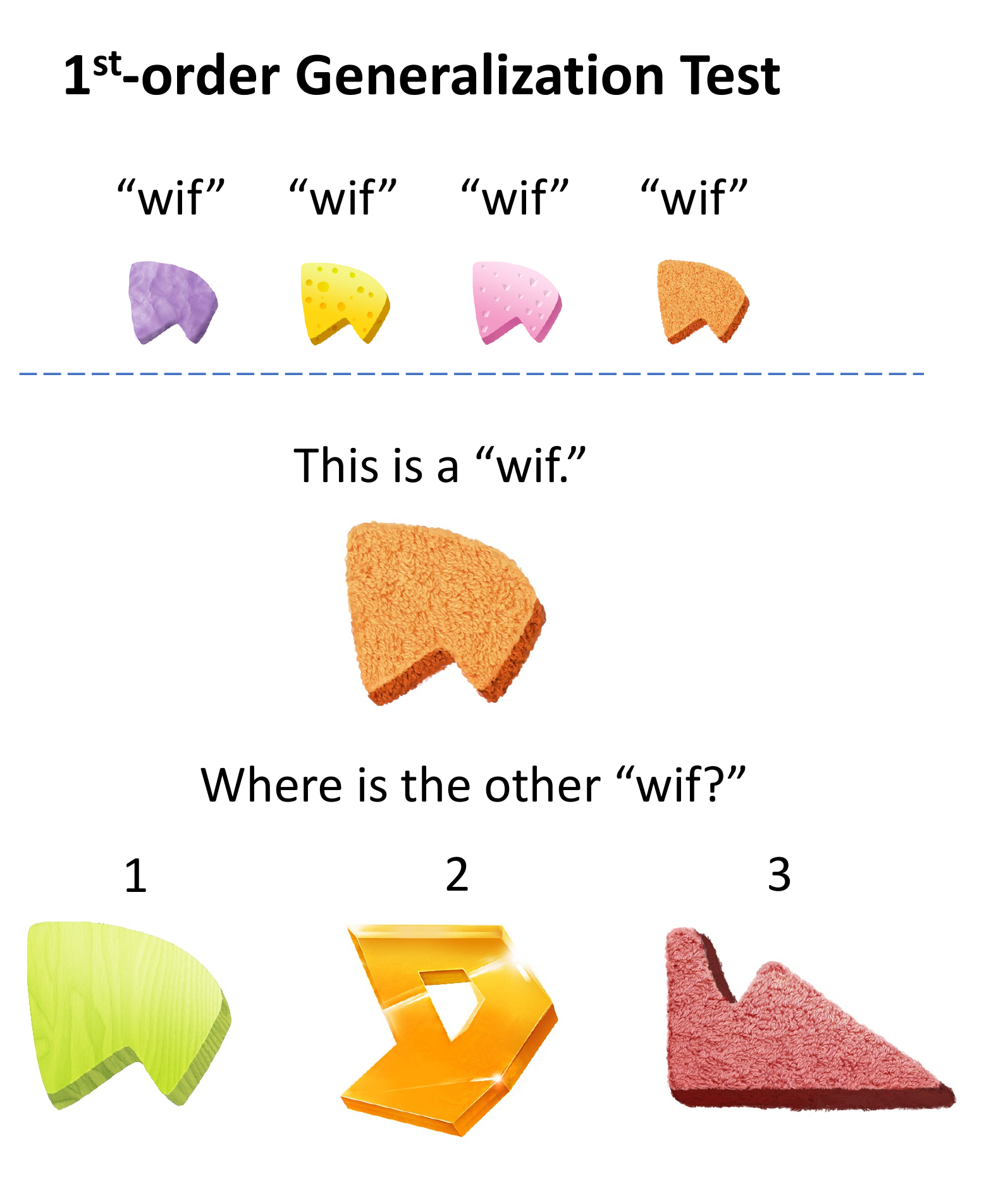}
            \end{center}
            \vspace*{-0.5em}
        	\caption{}
            \label{fig:gen_test_o1}
        \end{subfigure}
        \begin{subfigure}[b]{0.225\textwidth}
            \begin{center}
                \includegraphics[width=\linewidth]{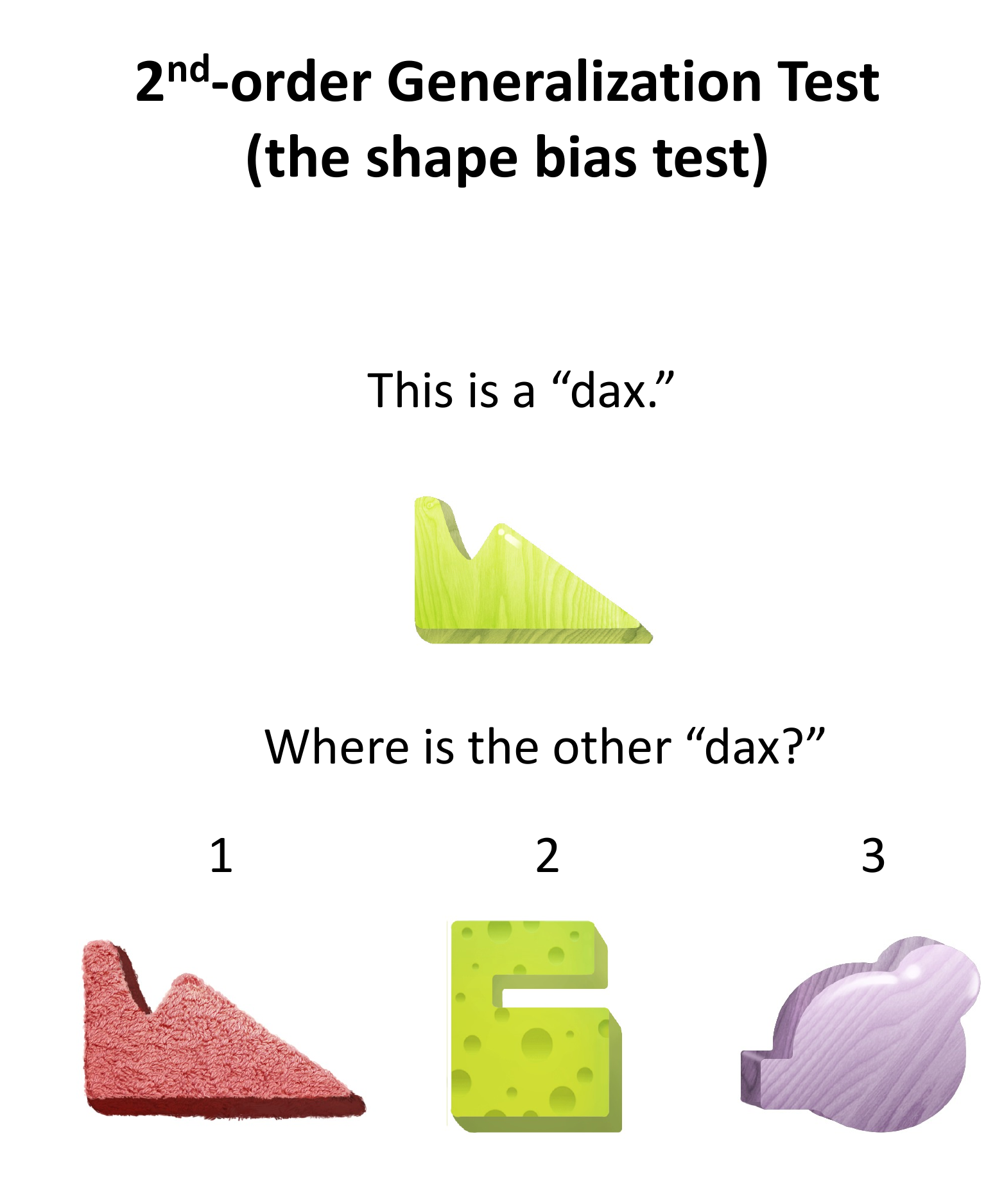}
            \end{center}
            \vspace*{-0.5em}
            \caption{}
            \label{fig:gen_test_o2}
        \end{subfigure}
    \end{center}
    \vspace*{-0.8em}
	\caption{Shape bias generalization tests. The 1st-order test, shown in (a), assesses if a child has learned to generalize a familiar object name to a novel exemplar according to shape. This is the first step of shape bias development. The 2nd-order test, shown in (b), assesses if the child has learned to generalize a novel name to a novel exemplar by shape, the second and final step of shape bias development.}
    \label{fig:gen_tests}
    \vspace*{-1.4em}
\end{figure}

If a learner extrapolates beyond the data, then another source of information must make up the difference; prior knowledge or ``inductive biases" must help constrain the space of models considered by the learner \citep{Tenenbaum2011, Mitchell2013, Lake2017}. For example, children make
use of the shape bias--the assumption that objects that have the same name will
tend to have the same shape--when learning new object names, and thus they
attend to shape more often than color, material and other properties when
generalizing a novel name to new examples (Fig. \ref{fig:gen_test_o2}) \citep{Landau1988}.
Similarly, children assume that object names are mutually exclusive, i.e. that
a novel name probably refers to a novel object rather than a familiar object
\citep{Markman1988}. Although the origin of inductive biases is not always
clear, results show that children, adults and primates can ``learn-to-learn" or
form higher-order generalizations that improve the efficiency of future
learning \citep{Harlow1949, Smith2002, Dewar2010}.

Researchers have proposed a number of computational models to explain
how inductive biases are acquired and harnessed for future learning.
Hierarchical Bayesian Models (HBMs) enable probabilistic inference at multiple
levels simultaneously, allowing the model to learn the structure of individual
concepts while also learning about the structure of concepts in
general \citep{Gelman2013, Kemp2007, Salakhutdinov2012}. These models have been used to
explain various forms of ``learning-to-learn," including learning a shape bias
\citep{Kemp2007}. However, it is currently difficult to apply HBMs to the type of high-dimensional visual and auditory stimuli that children receive; there have been successes \citep{Salakhutdinov2013,Lake2015}, but neural networks are still the most general solution to learning effectively from many different forms of raw data \citep{LeCun2015}.
Utilizing this property, here we use neural networks to study learning-to-learn in different settings of varying stimulus complexity, with the goal of isolating the fundamentals of the learning dynamics.

Most related to our work here are studies by \cite{Colunga2005} and
\cite{Ritter2017} investigating neural network accounts of shape bias
development. \cite{Colunga2005} showed that a simple recurrent neural network,
trained via Hebbian learning, can acquire a shape bias for solid objects and a material bias for non-solid objects. These simulations demonstrate that neural networks can form different expectations for different kinds of objects, but they raise many new
questions regarding the conditions required to develop
these types of biases. For example, the authors used highly simplified bit-vector data,
and it is unclear whether their findings generalize to more complex or realistic stimuli.
Furthermore, the authors did not systematically vary the quantity of experience provided to the networks, and thus we do not know the exact conditions in which biases arise and whether these networks can compete with the strong sample efficiency of HBMs \citep{Kemp2007}.
In a recent study,
\cite{Ritter2017} found that performance-optimized deep neural networks (DNNs) develop the shape bias when trained on the popular ImageNet object recognition dataset consisting of raw
naturalistic images. These results highlight an exciting possible
connection between large-scale DNNs and developmental psychology, though
many questions still remain. ImageNet--which contains about 1200 labeled examples of 1000 different
object categories--is a poor proxy for the experience of a developing child, who
typically develops a shape bias with no more than 50-100 object words
in her vocabulary \citep{GershkoffStowe2004}. Whether
these networks can acquire the shape bias with more appropriate training sets is unclear. Furthermore, although the development of the shape bias is known to predict the onset of vocabulary acceleration in children
\citep{GershkoffStowe2004}, we do not know whether the same holds for DNNs.

In a related study, \cite{Hill2017} trained a neural network agent to navigate around a virtual 3D world and collect objects according to name-based language commands. Although the authors draw inspiration from developmental psychology, the agent in this experiment is asked to learn a variety of tasks simultaneously: visual perception, language comprehension and navigation.
Further work is necessary to isolate the dynamics of learning-to-learn in neural networks.

We investigate the development and influence of inductive biases in neural
networks using artificial object stimuli that allow us to systematically vary
the quality and form of the experience provided. Specifically, we use an experimental paradigm from developmental psychology \citep{Smith2002} to train and evaluate the networks.
Beginning with simple bit-vector data akin to \cite{Colunga2005}, we systematically vary
the number of categories and the number of examples in the training set, and for each pairing the trained networks are evaluated for two different forms of generalization (Fig. \ref{fig:gen_tests}) as well as for changes in perceptual sensitivity. Parallel experiments are then performed with raw image data, where each image consists of a 2D object with a particular shape, color and texture.
In a final set of experiments, we examine the dynamics of learning-to-learn by analyzing the relationship between shape bias acquisition and the rate of word learning, mirroring an analogous study from developmental psychology \citep{GershkoffStowe2004}.
\vspace*{-0.5em}

\section{Experiments}
\label{sec:experimental_paradigm}

\begin{figure*}[t]
    \begin{center}
        \begin{subfigure}[b]{0.48\textwidth}
            \begin{center}
                \includegraphics[width=0.98\textwidth]
                {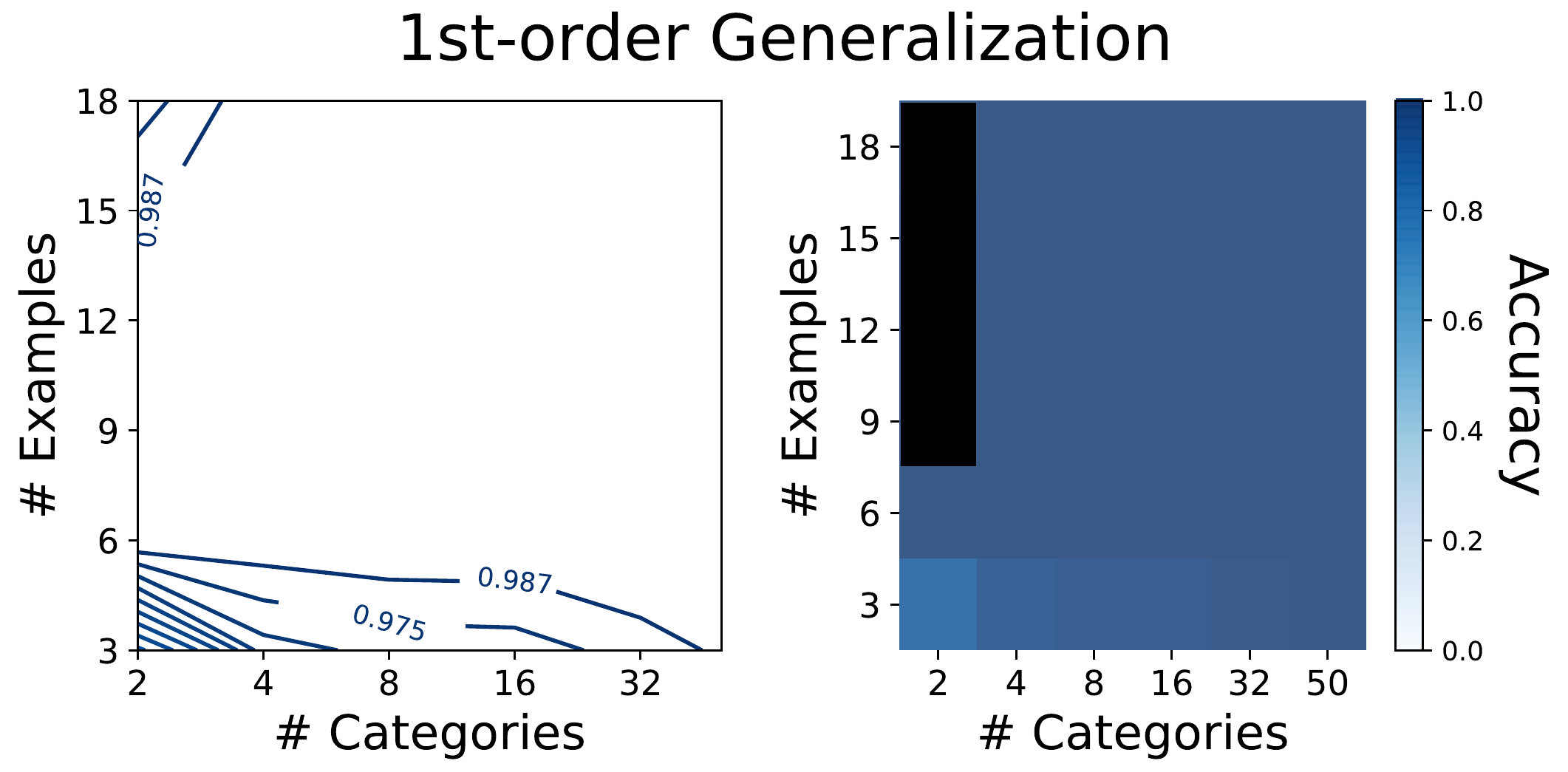}
            \end{center}
        \end{subfigure}
        \begin{subfigure}[b]{0.48\textwidth}
            \begin{center}
                \includegraphics[width=0.98\textwidth]
                {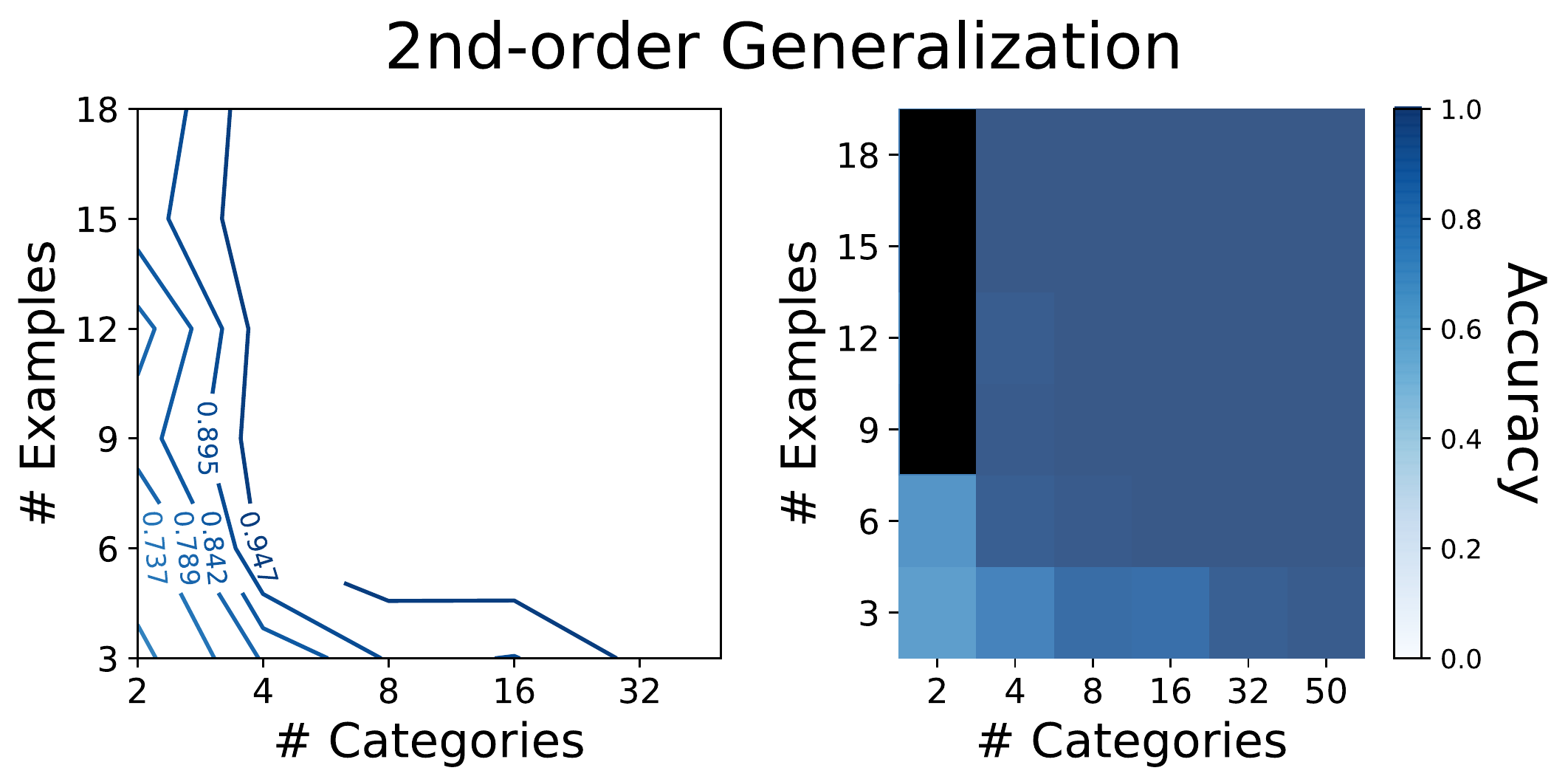}
            \end{center}
        \end{subfigure}
    \end{center}
    \vspace*{-0.8em}
    \caption{MLP generalization results for explicit shape bias training with various training set sizes.
    The number of categories and number of examples per category provided to the network are shown on the x and y axes, respectively. Plots show accuracy over 1000 trials of the specified generalization test, averaged from 10 training runs. The same data is shown in both contour and heatmap format. With 2 categories, only 8 unique examples are feasible; thus, N/A results are blacked out.}
    \label{fig:mlp_gen_results}
    \vspace*{-0.5em}
\end{figure*}

We set out to train neural networks with a learning paradigm used to guide toddlers to the shape bias \citep{Smith2002}.
In this paradigm, the learner acquires new object names that are organized exclusively by shape, such that different instances of the same object category are identical in shape but contrast sharply in color and material.
This is reflective of the fact that a child's early noun vocabulary consists predominantly of shape-based categories \citep{Samuelson1999}, although not with the same purity as
the shape bias training.
As in previous computational modeling work \citep{Kemp2007,Colunga2005}, we focus on purified training with shape-based categories,
since it provides a controlled
test of the artificial learner's ability to make higher-order generalizations across varying quantities of training experience.

In \cite{Smith2002}, 17-month-old children were taught 4 new object names (``wif'', ``dax'', etc.) over 7 weeks via weekly play sessions. Objects in the study were 3D formations constructed of
various materials; each object contained a specific shape, color and texture (material), and their names were organized strictly by shape.
During weekly sessions, children played with each object while an adult announced
its name
repeatedly. By the end of the study, the children had acquired a shape bias--i.e., they had formed the inductive bias that a novel name should be generalized by shape as opposed to color or texture. A control group of children who did not partake in the play sessions did not form this bias.

We use the training paradigm of \cite{Smith2002} to study inductive bias learning in neural networks with artificial object datasets. We first perform our computational experiments with abstract bit-vector stimuli, followed by experiments with
raw image data.
Each constructed object is assigned a shape, color and texture. We train simple neural networks to
label objects with category names based on shape. To understand the necessary conditions
for successful inductive bias learning, training is performed with various dataset sizes, varying both the number of categories and the number of examples of each category provided to the network.
We evaluate the generalization capabilities of the network for each training
set using 2 generalization tests modeled after the 2 tests of
\cite{Smith2002}, depicted in Fig. \ref{fig:gen_tests}.

\subsubsection{\nth{1}-order generalization test.} For this
test, toddlers are first presented with an exemplar object
that they have seen during training (``wif'' in Fig. \ref{fig:gen_tests}a).
Then, they are presented with 3 test objects
that they have not seen: 1 that matches the exemplar in shape (item 1 in Fig. \ref{fig:gen_tests}a), 1
that matches in color (item 2), and 1 that matches in texture (item 3). For each potential match,
the other 2 stimulus attributes are novel. The toddlers are asked to select
which of the 3 test objects share the same name as the exemplar.
Performance is measured as the fraction of trials in which the child
selected the correct object, i.e. the shape match.
To simulate this test, we create an evaluation set containing groupings of 4 sample objects: an exemplar, a shape match, a color match, and a texture match.
The activations of our network's hidden layer are obtained in response to each object. We then evaluate the cosine similarity
between the activations of the exemplar and each test object to determine which object the network perceives to be most similar.
Accuracy is defined as
the fraction of groupings for which the correct (shape-similar)
object is chosen.

\subsubsection{\nth{2}-order generalization test.} For this
test, toddlers are first presented with an exemplar object that has a novel label (e.g., ``teema'')
as well as a novel shape, color and texture. From there, the trial proceeds similarly to those of the
\nth{1}-order: a shape match, color match and texture match are presented,
and the child must select which test object she believes to share a name with the
exemplar. All shapes, colors and textures are novel to the child in this test.
We simulate the \nth{2}-order test with artificial object stimuli similarly to the
\nth{1}-order case, again using last hidden layer features to evaluate perceptual similarity.

In all simulations, we record accuracy over 1000 simulated test trials as the performance metric for each generalization.
\subsection{Experiment 1: Multilayer perceptron trained on synthetic objects}
\label{sec:simple_mlp}

\begin{figure}[t!]
    \begin{center}
        \includegraphics[width=0.4\textwidth]{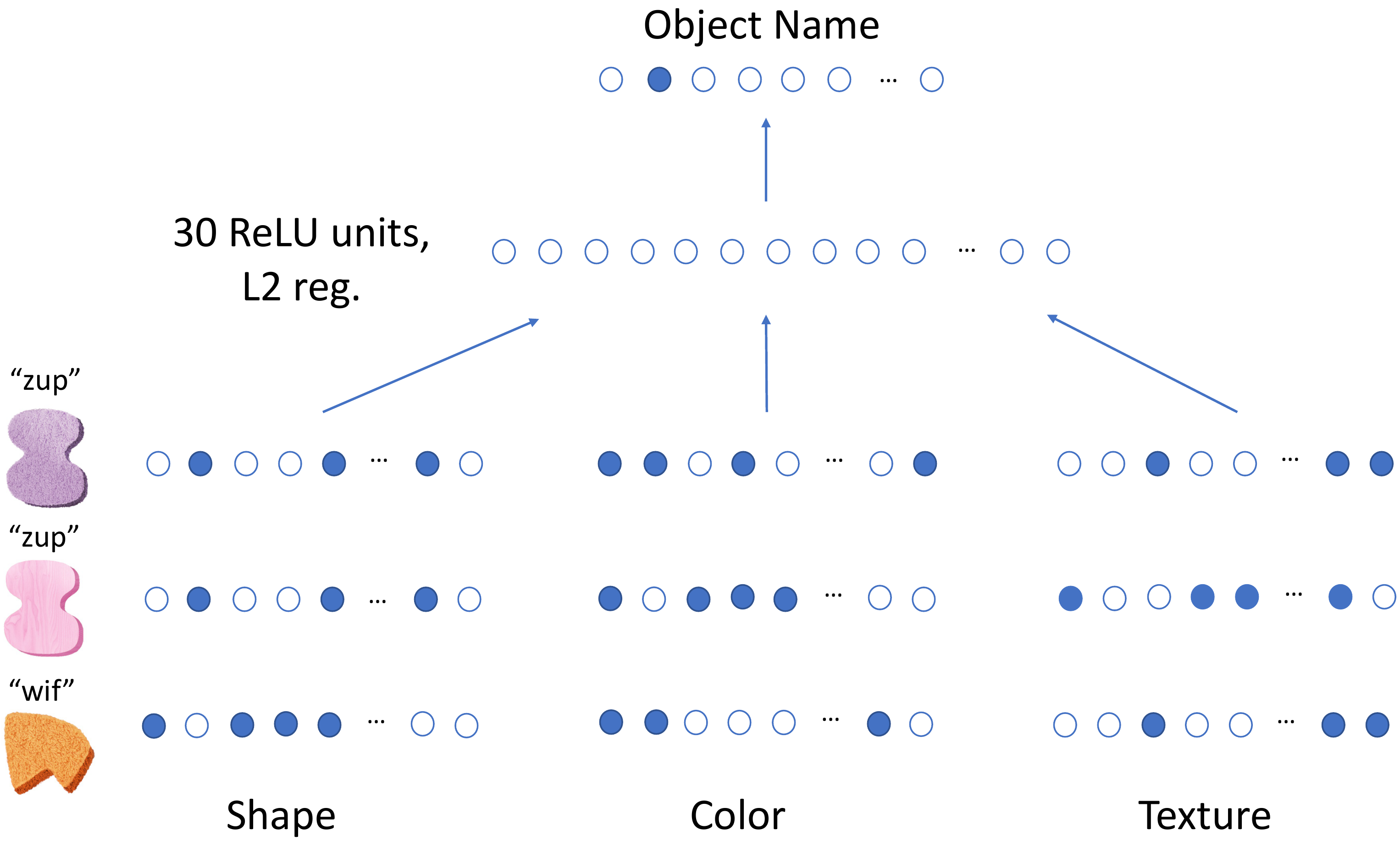}
    \end{center}
    \caption{Multilayer perceptron architecture. Shape, color and texture attribute vectors are concatenated and fed to a 30-unit hidden layer, followed by a classification layer. 3 example input objects are shown (only one is presented at a time to the network).}
    \label{fig:mlp_diagram}
    \vspace*{-1.2em}
\end{figure}

Our first experiment aims to study inductive bias learning in its purest form, using
synthetic stimuli with maximal control. Objects are abstract binary patterns, divided into 3 input pools of 20 binary units each (representing the shape, color and texture of the objects; see Fig. \ref{fig:mlp_diagram}).
We varied the number of categories and number of examples per category in the training set. For datasets with $N$ categories and $K$ examples, we randomly generate $N$ shape patterns, $N$ color patterns, and $N$ texture patterns. For all 3 attributes, each pattern is replicated $K$ times, ensuring equal entropy across the 3. The shape patterns are then aligned with object labels, and the remaining 2 attributes are permuted randomly to create the dataset. A holdout set of shapes, colors and textures is retained for the generalization tests.

We train a multilayer perceptron (MLP) to name objects, as shown in Fig. \ref{fig:mlp_diagram}. The network has an input layer of 60 units, a hidden layer of 30 rectified linear units (ReLUs) with L2 regularization,
and a softmax output layer to classify the object by name. The softmax layer has $N$ units (1 for each label).
We train the network for 200 epochs using negative log-likelihood loss, RMSProp, and batch size min(32, $\frac{N*K}{5}$).\footnote{For details about the selection of architectures and training parameters in Experiments 1 \& 2, see Supplemental Material (\hyperref[sec:sm1]{SM 1}).}

\subsubsection{Results.} Initially, as would be expected given the data format, shape is treated the same as other attributes. In the \nth{2}-order generalization test, a randomly initialized network selects test objects with the following ratios, on average (50 trials): shape 35\%, color 33\% and texture 32\%.
We then trained the network with various dataset sizes. Results for the \nth{1}- and \nth{2}-order generalizations are shown in Fig. \ref{fig:mlp_gen_results}, where each setup is an average over 10 networks with different random seeds. We note that acquisition of the \nth{1}-order generalization requires less data than that of the \nth{2}-order, as predicted by the 2-phase hypothesis \citep{Smith2002}. Success in the \nth{1}-order test indicates that the network is learning successfully and generalizing to new examples of the training classes. Networks that achieve an accuracy of 0.7 or higher on the \nth{2}-order test show a substantial shape bias, and the MLP reaches this threshold at the following points: $N$=2 \& $K$=6 (accuracy 0.71) and $N$=4 \& $K$=3 (accuracy 0.80). These results reproduce the general pattern of the Hierarchical Bayesian Model (HBM) in \cite{Kemp2007} and toddlers in \cite{Smith2002}, who neared the 0.7 shape bias threshold with $N$=4 \& $K$=2 (although the toddlers also receive external experience). In contrast, \cite{Colunga2005} used $N$=10 \& $K$=100 to obtain the shape bias in their networks, using similar abstract patterns. Although HBMs are often noted for their data efficiency, in this case, the neural network was competitive for making \nth{2}-order generalizations from limited data.

\begin{figure}[t!]
    	\begin{subfigure}[b]{0.235\textwidth}
            \begin{center}
                \includegraphics[width=\linewidth]{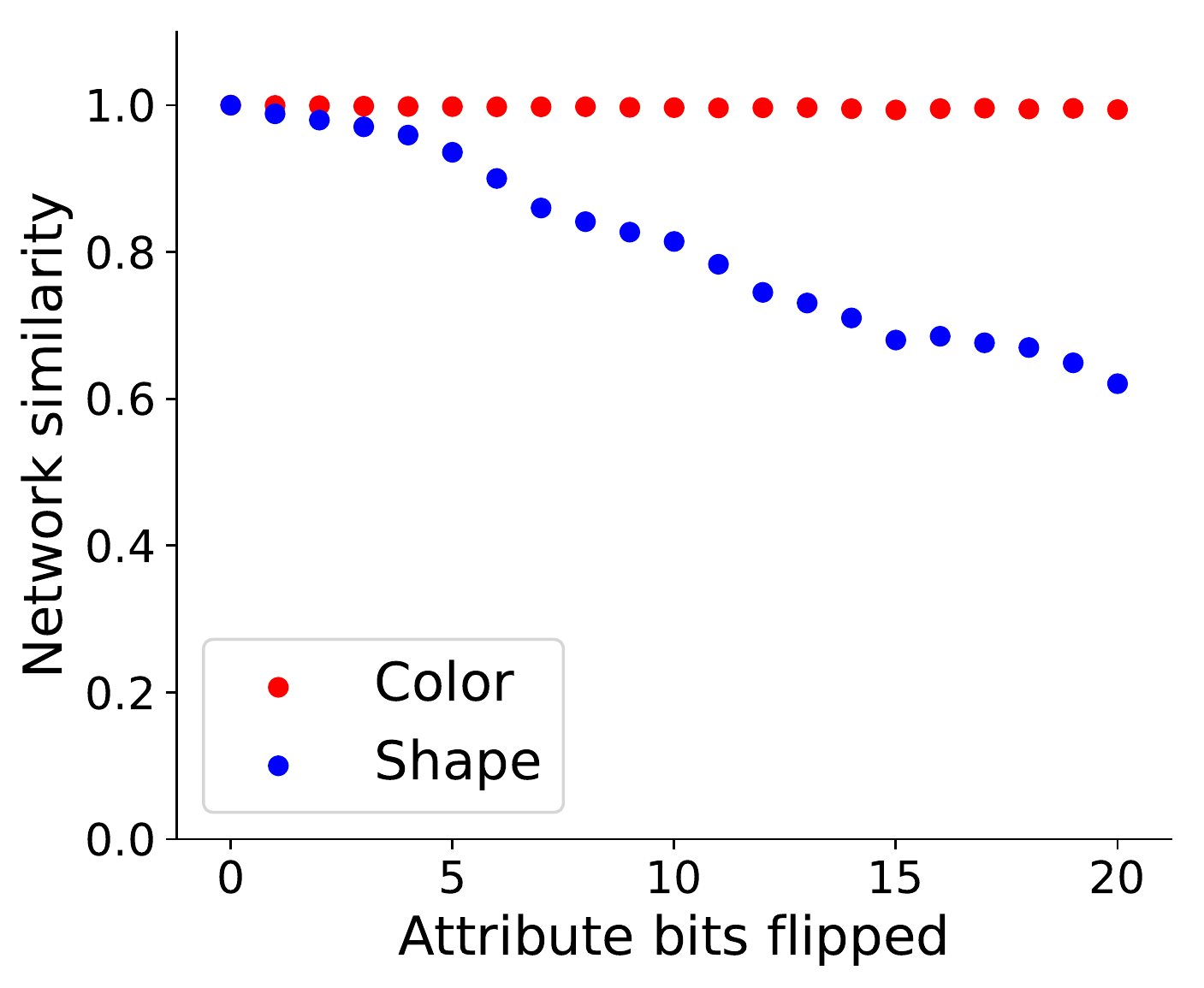}
            \end{center}
            \vspace*{-0.5em}
        	\caption{MLP}
            \label{fig:mlp_parametric}
        \end{subfigure}
        \begin{subfigure}[b]{0.235\textwidth}
            \begin{center}
                \includegraphics[width=\linewidth]{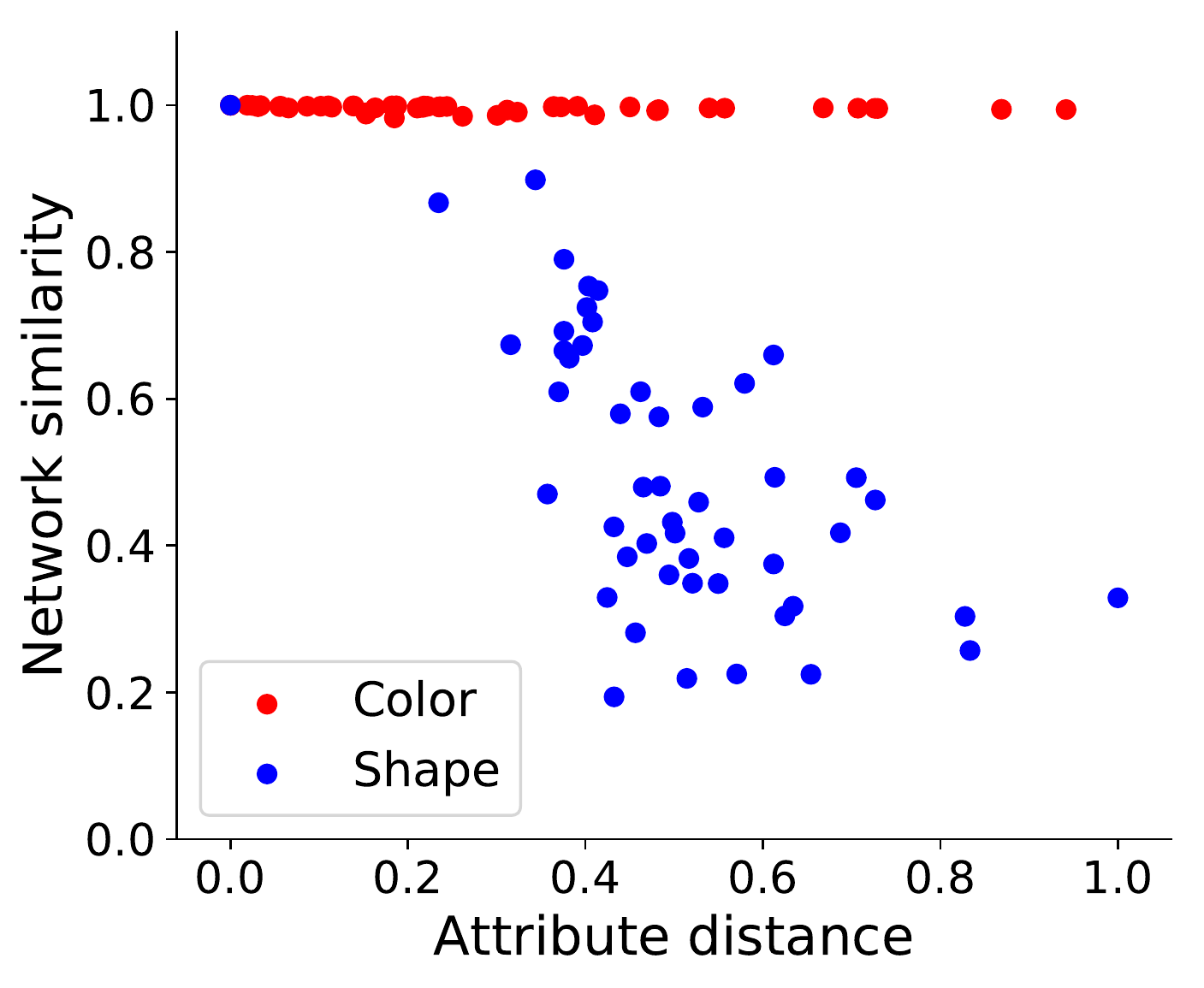}
            \end{center}
            \vspace*{-0.5em}
            \caption{CNN}
            \label{fig:cnn_parametric}
        \end{subfigure}
        \begin{subfigure}[b]{0.5\textwidth}
            \begin{center}
                \includegraphics[width=0.7\linewidth]{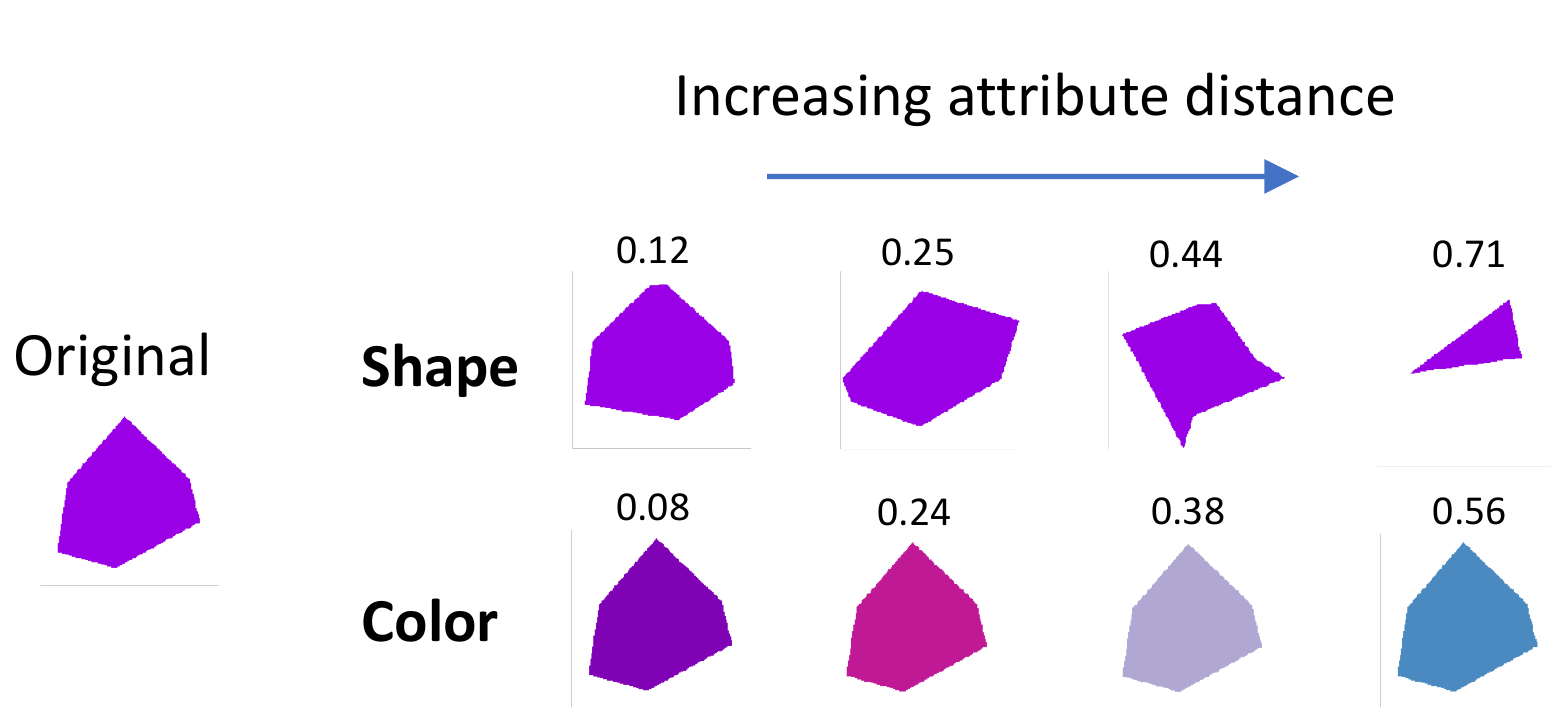}
            \end{center}
            \caption{Distance along stimulus dimensions}
            \label{fig:parametric_distance}
        \end{subfigure}
    \vspace*{-1.5em}
    \caption{Perceptual (network) similarity as a function of physical (attribute) distance. A test stimulus is systematically altered along either its shape or color dimension. Network similarity scores are computed between the original stimulus and its altered counterpart.}
    \label{fig:parametric_tests}
    \vspace*{-1.0em}
\end{figure}

As another way of demonstrating the learned sensitivity to shape, we perform parametric manipulations of the stimuli. Using an MLP trained with $N$=4 \& $K$=6, we probe the shape bias by selecting a novel test stimulus and systematically flipping bits, recording the network similarity between the modified stimulus and the original. For comparison, a similar test is also performed with color. Results are shown in Fig. \ref{fig:mlp_parametric} for 1 test stimulus. Clearly, the network is far more sensitive to changes in shape than changes in color.
\subsection{Experiment 2: Convolutional network trained on synthetic objects}
\label{sec:simple_cnn}

Our first experiment used highly simplified training stimuli for maximal experimental control. One strength
of modern neural network architectures is that they can learn effectively
from data in raw and complex forms, a fact we take advantage of in developing
Experiment 2. Here we ask whether similar learning-to-learn results can be achieved using synthetic object stimuli presented as raw images. This setup presents a more challenging learning problem for the neural network, in terms of making both \nth{1}- and \nth{2}-order generalizations, since understanding shape requires making abstractions that go substantially beyond separating a pool of input units that directly encode the dimension, as in Experiment 1.

The stimuli are constructed as follows.  Each object is a 2D shape of a specified color placed over white background (200x200).
Texture is represented in a fourth image channel, independent of RGB space.\footnote{This design choice was made to avoid an initial shape bias; with texture overlaid in RGB space, a randomly initialized network exhibits the shape bias. Furthermore, the participants in \cite{Smith2002} physically touch each object during play, indicating that they have access to additional non-visual information.}

\begin{figure}[t!]
	\vspace*{-0.4em}
    \begin{center}
    	\begin{subfigure}[b]{0.2\textwidth}
        	\includegraphics[width=\linewidth]{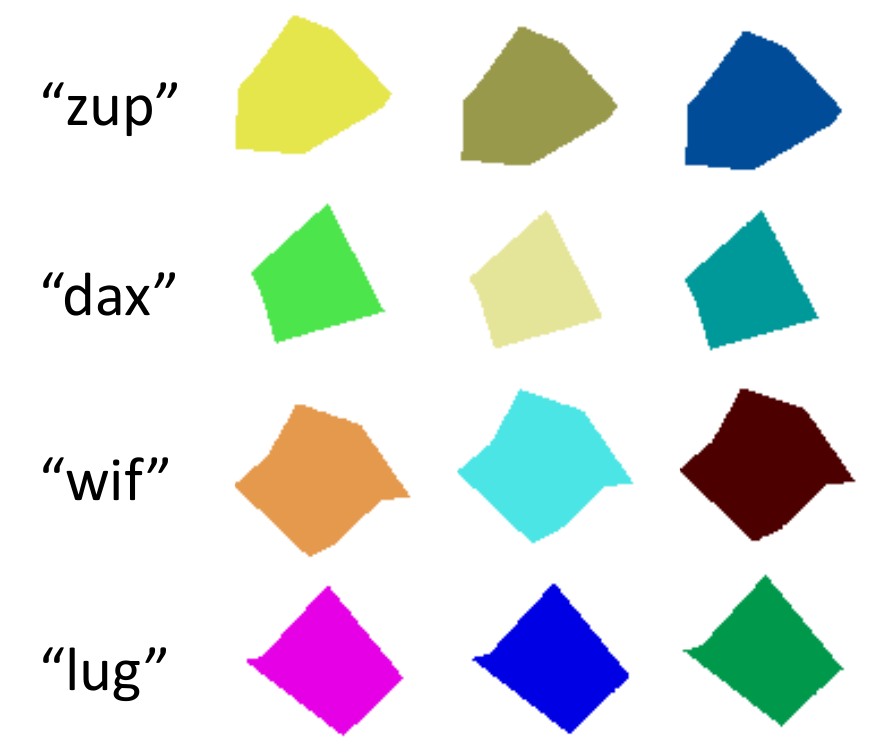}
            \caption{}
        \end{subfigure}
        \begin{subfigure}[b]{0.17\textwidth}
        	\includegraphics[width=\linewidth]{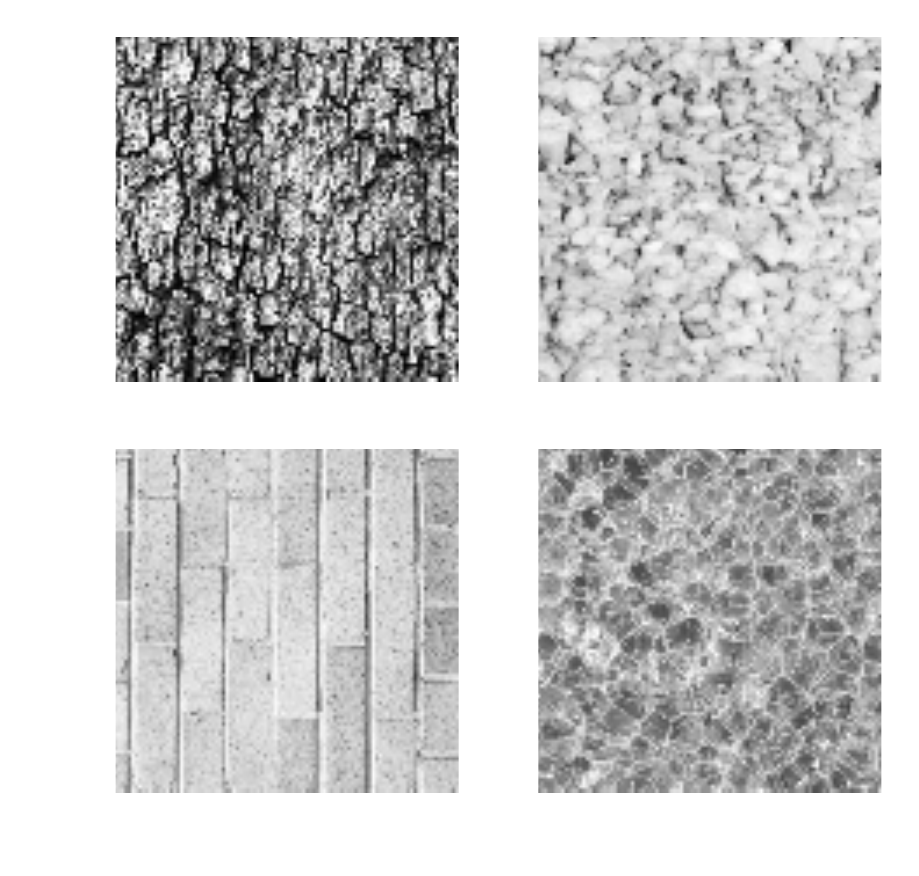}
            \caption{}
        \end{subfigure}
    \end{center}
    \vspace*{-0.8em}
    \caption{Training stimuli for Experiment 2. (a) novel objects with various shapes and colors (the first 3 input channels). (b) a few examples of textures that might be found in the 4th input channel.}
    \label{fig:generated_images}
    \vspace*{-1.2em}
\end{figure}

\begin{figure*}[t!]
    \begin{center}
        \begin{subfigure}[b]{0.48\textwidth}
            \begin{center}
                \includegraphics[width=0.98\textwidth]{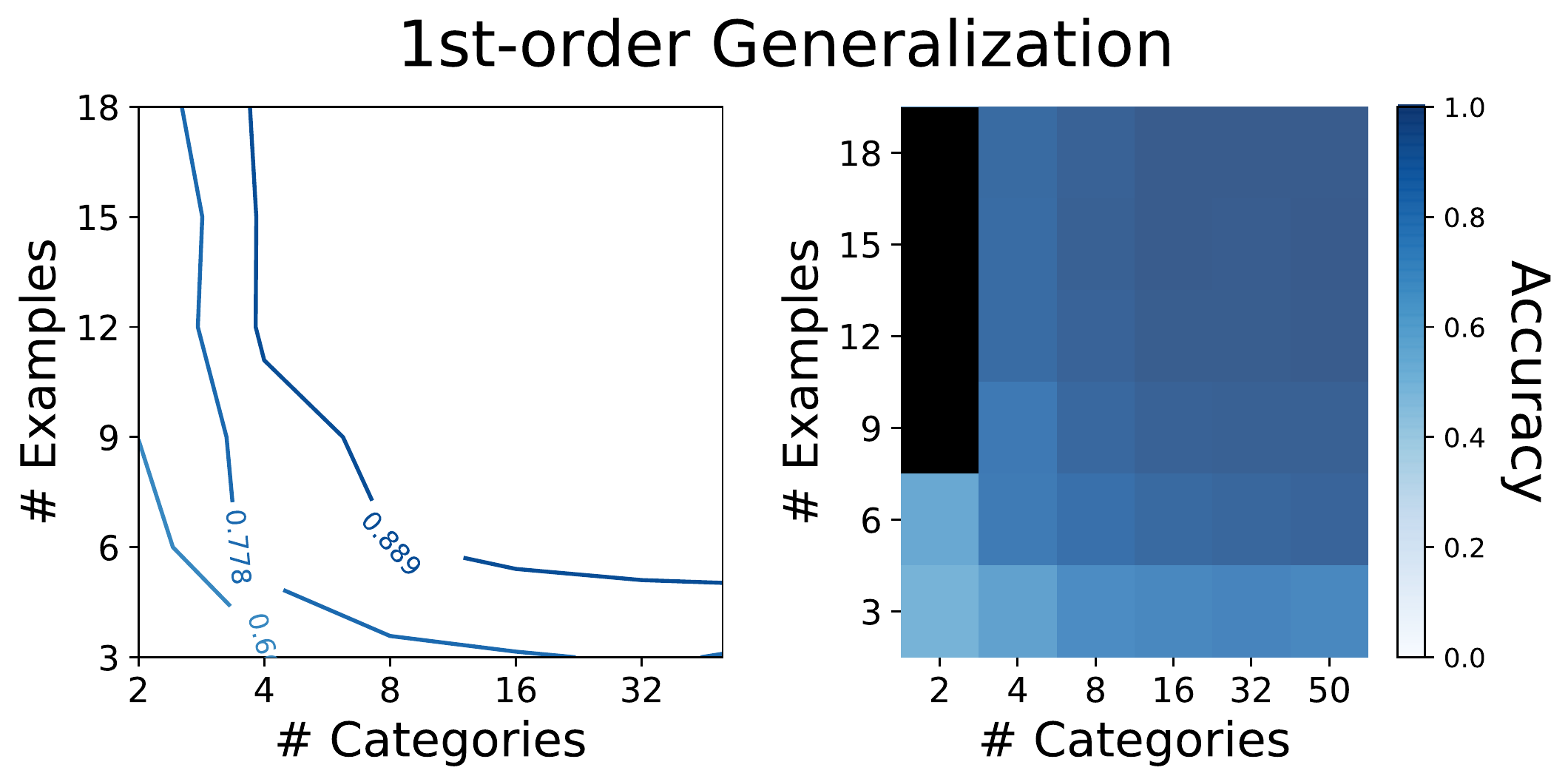}
            \end{center}
        \end{subfigure}
        \begin{subfigure}[b]{0.48\textwidth}
            \begin{center}
                \includegraphics[width=0.98\textwidth]{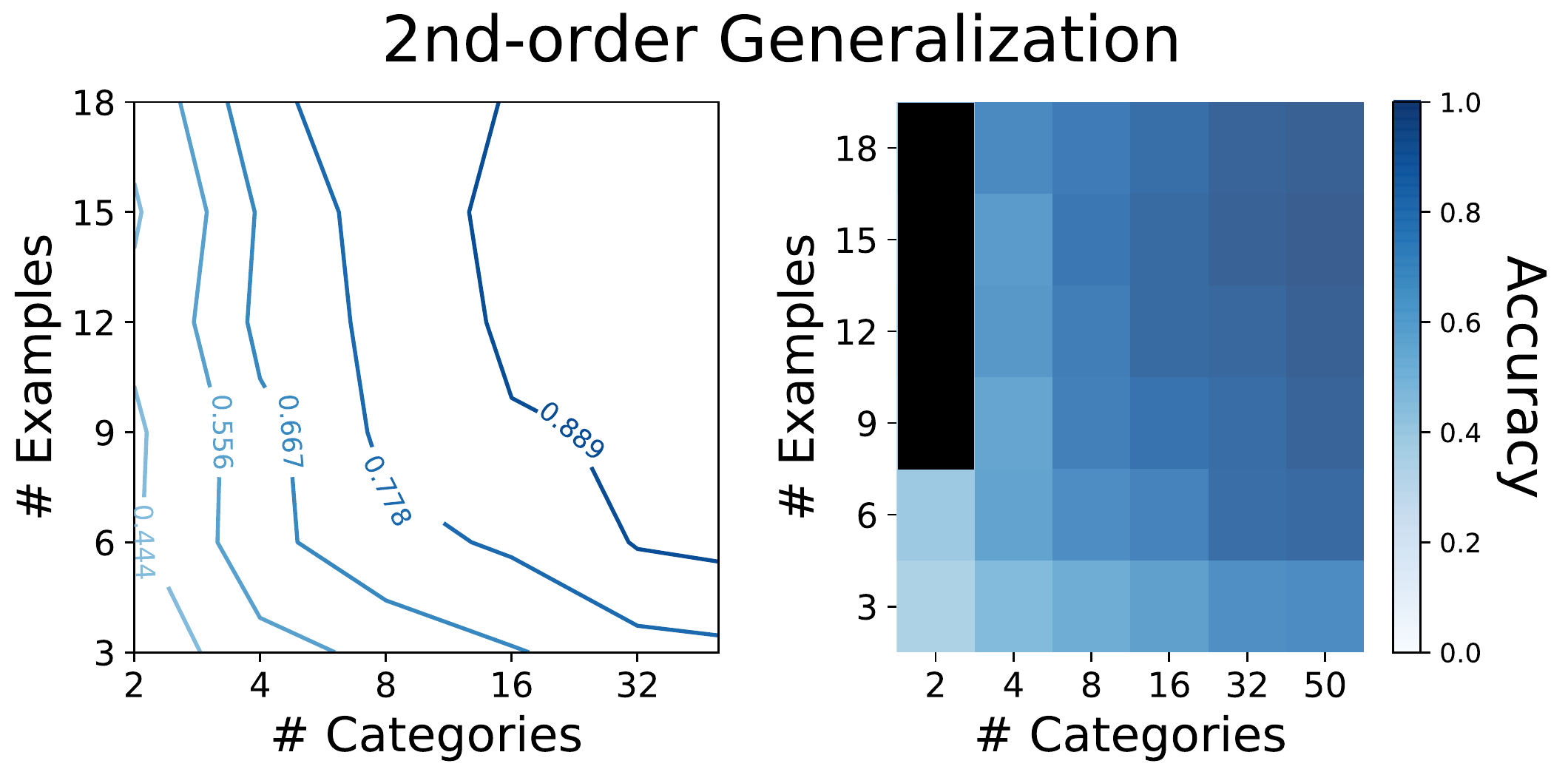}
            \end{center}
        \end{subfigure}
    \end{center}
    \vspace*{-0.8em}
    \caption{CNN generalization results for various training set sizes. Results show the average of 10 training runs. See Fig. \ref{fig:mlp_gen_results} for details.}
    \vspace*{-0.8em}
    \label{fig:cnn_gen_results}
\end{figure*}

Examples of our objects are shown in Fig. \ref{fig:generated_images}. Object shapes are polygons of random order (uniform 3-10) and randomly sampled vertices, with preference given to points near image boundaries in order to ensure visible-sized objects. Colors are generated to span the RGB vector space with even separation. We use black and white textures from the Brodatz database \citep{Brodatz1966}
for our texture categories. A holdout set of shapes, colors and textures is again retained for testing.

\begin{figure}[b!]
	\vspace*{-0.7em}
    \begin{center}
        \includegraphics[width=0.47\textwidth]{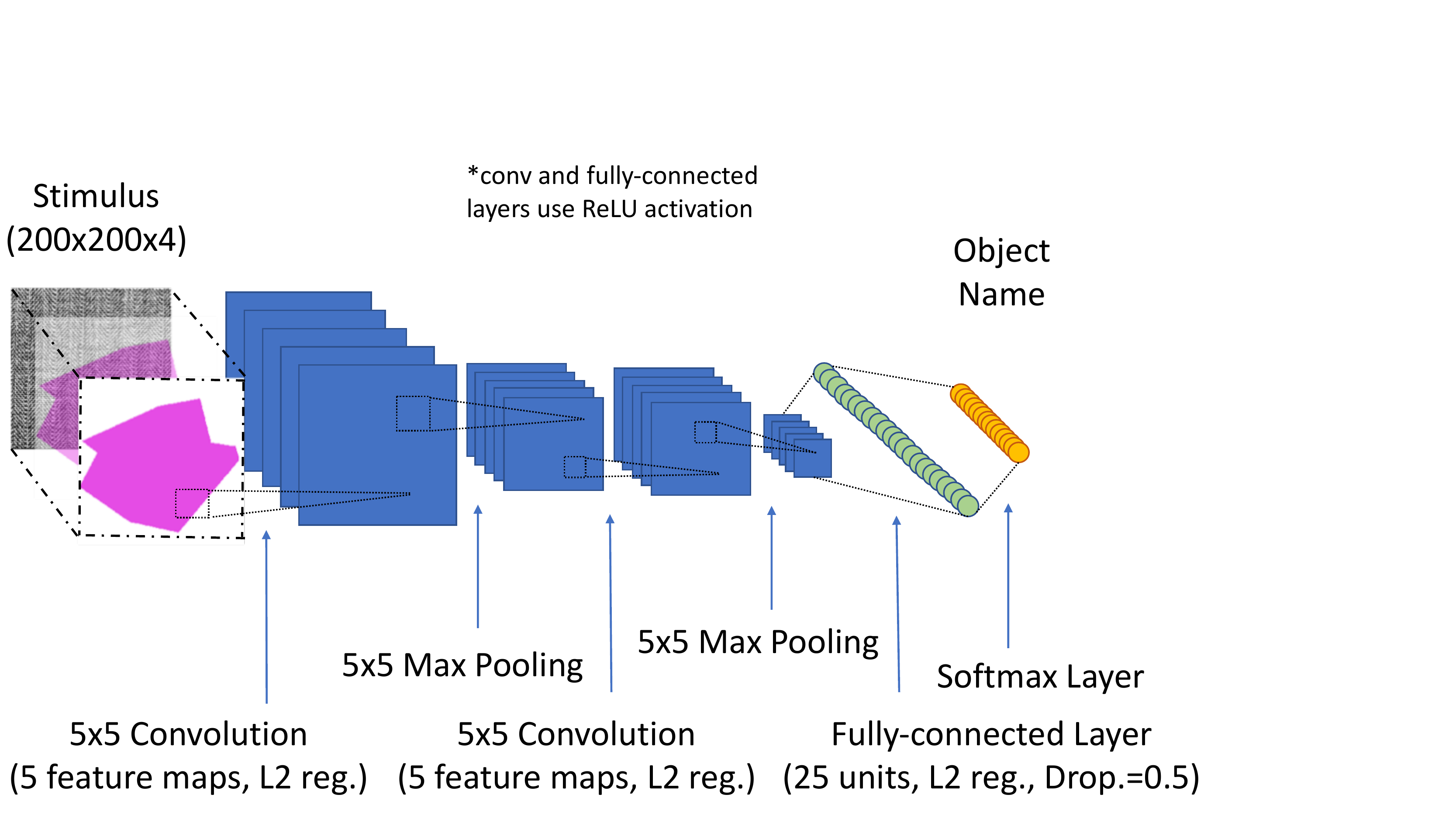}
    \end{center}
    \vspace*{-0.5em}
    \caption{Convolutional network architecture. The network receives 4D image stimuli and is trained to label the object in the image with a category name that is based on shape.}
    \label{fig:cnn_diagram}
\end{figure}

We train a multi-layer convolutional neural network (CNN) \citep{LeCun2015} consisting of two convolution layers with five feature maps, each followed by a max pooling
layer. A depiction of this architecture is shown in Fig.
\ref{fig:cnn_diagram}. The last pooling layer is followed by a fully-connected layer of 25 ReLU
units, and the softmax layer again varies in size according to the number of
categories.
Both the convolutional layers and the fully-connected layer use L2 regularization, the latter also with dropout=0.5. Each object is randomly shifted around image space by a small offset (train and test alike). Training details mimic the MLP, but with 400 epochs.

\subsubsection{Results.} The randomly initialized network makes \nth{2}-order selections with the following ratios: shape 38\%, color 42\% and texture 20\%. We trained the network using varying dataset sizes, as with our MLP.
Results are shown in Fig. \ref{fig:cnn_gen_results}.
Similarly to the MLP, acquisition of the \nth{1}-order generalization requires less data than that of the \nth{2}-order, supporting the notion that learning the training classes is a simpler task than forming higher-order generalizations.
Using the same shape bias threshold of 0.7 \nth{2}-order score, we find a number of important transition points: $N$=32 \& $K$=3 (accuracy 0.74), $N$=8 \& $K$=6 (accuracy 0.75), and $N$=4 \& $K$=12 (accuracy 0.70). The CNN is thus capable of learning a shape bias from as few as 6 examples of 8 categories, a significant feat given the scale of the input. Notably, the network is able to learn this bias with much less data than \cite{Colunga2005} using a data form that is significantly more complex. The CNN of \cite{Ritter2017} used roughly $N$=1000 \& $K$=1200, and developed a shape bias of 0.68 on a shape and color-only task. A key takeaway from our results is that, with concentrated training effort, it is possible to learn this bias from much less data using high-dimensional color images.

As in Experiment 1, we also parametrically manipulated the stimuli to analyze the network's sensitivity to changes along different stimulus dimensions, finding strong sensitivity to shape but not color (Fig. \ref{fig:cnn_parametric} \& \ref{fig:parametric_distance}; see \hyperref[sec:sm2]{SM 2} for details). Moreover, in additional experiments, we found that networks trained on color-based categories develop a color bias, and that this bias requires less data than the shape bias (see \hyperref[sec:sm3]{SM 3} for details).
\subsection{Experiment 3: The onset of vocabulary acceleration}
\label{sec:vocab_accel}
Our previous experiments confirm that simple neural networks can develop the shape
bias from a relatively small number of categories and examples.
It remains unclear, however, how the dynamics of bias acquisition relate to the dynamics of word learning. \cite{GershkoffStowe2004}
showed that the development of the shape bias in toddlers predicts the onset of vocabulary acceleration during early word learning, a phase that begins at ages 16-20 months. Studying 8 children during regular lab sessions at 3-week intervals, the authors found that increasing attention to shape was correlated with increasing rate of vocabulary acquisition in participants. Fig. \ref{fig:learning_curves_children} shows the individual growth curves of vocabulary size and shape response for each child. The former variable is measured as the cumulative number of nouns in the child's vocabulary, and the latter as the cumulative number of times that the child has selected the shape match in a shape bias task akin to the \nth{2}-order test. Although the vocabulary curve shows cumulative nouns in whole, the authors also recorded cumulative ``count nouns" for each participant, a subset of nouns that is well organized by shape. We focus on the statistics reported for count nouns, as this subset reflects the type of vocabulary that is influenced by the shape bias.


The authors found a few interesting correlations: {\bf1)} a correlation between increase in cumulative shape choices and increase in cumulative count nouns across sessions for an individual participant, averaged over participants [average $r=.75$; $p<.05$ for each], and {\bf2)} a correlation between average increase in shape choices over the whole experiment and average increase in count nouns, computed across participants [$r=.81$; $p<.05$].

\subsubsection{Methods.} Inspired by this study, we train a CNN using our raw image data with the goal of evaluating related correlation metrics for our networks. The participants of \cite{GershkoffStowe2004} were not explicitly trained for the shape bias like those of \cite{Smith2002}; they received natural experience in a home setting, which may have included some words organized by attributes other than shape.
Therefore, we train our CNN to simultaneously label the object's name, which correlates with shape, as well as its color and texture names.
Thus, there are now 3 softmax layers in the CNN from Fig. \ref{fig:cnn_diagram}, each of which branches independently from the fully-connected layer (see SM 4 for network details).
The number of categories along each label dimension and the loss weight assigned to that dimension are determined according to the natural statistics of the early human lexicon \citep{Samuelson1999}.\footnote{Children are taught object, color and material names independently. Loss weighting provides a good analog to this for CNN training. Assigning a weight of 0.6 to object name labeling mirrors presenting this type of name 60\% of the time in training.
For details on loss weighting, see Supplemental Material (\hyperref[sec:sm4]{SM 4}).}
The chosen ratios are as follows: 60-20-20 shape-color-texture names (36, 12 and 12 categories, respectively). 10 examples of each shape are used, and colors and textures are assigned at random to each stimuli from their 12 categories.
We keep a cumulative count of the number of count nouns in the network's vocabulary, defined as the number of shape categories for which the network has achieved 80\% or greater accuracy on the training set. We also keep a cumulative count of shape choices the network makes in a 500-trial \nth{2}-order test. This process is repeated with 20 networks, using a different random seed for each network.

\subsubsection{Results.} 
\begin{figure}[t!]
	\begin{center}
        \begin{subfigure}[b]{0.5\textwidth}
            \begin{center}
                \includegraphics[width=\textwidth]{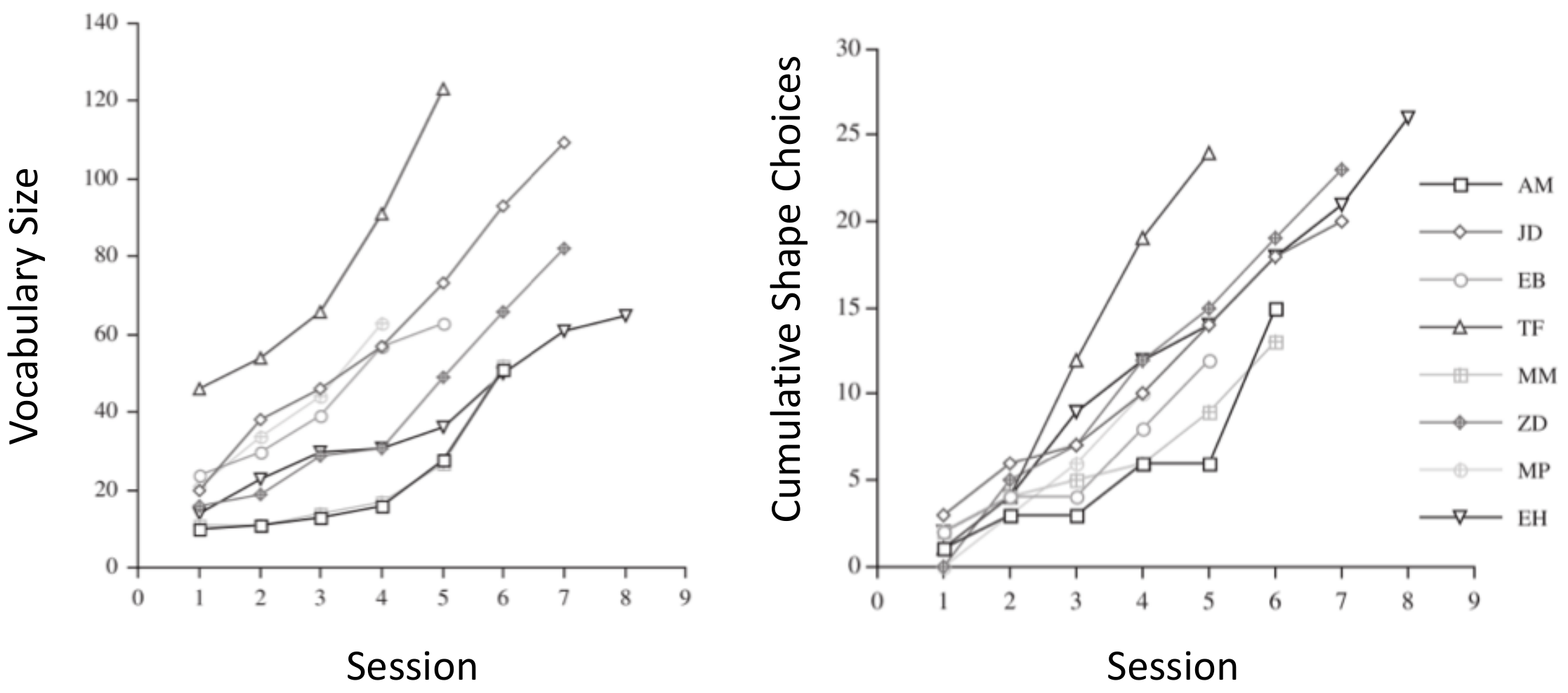}
            \end{center}
            \vspace*{-0.9em}
            \caption{Children}
            \vspace*{0.7em}
            \label{fig:learning_curves_children}
        \end{subfigure}
        \begin{subfigure}[b]{0.5\textwidth}
            \begin{center}
                \includegraphics[width=\textwidth]{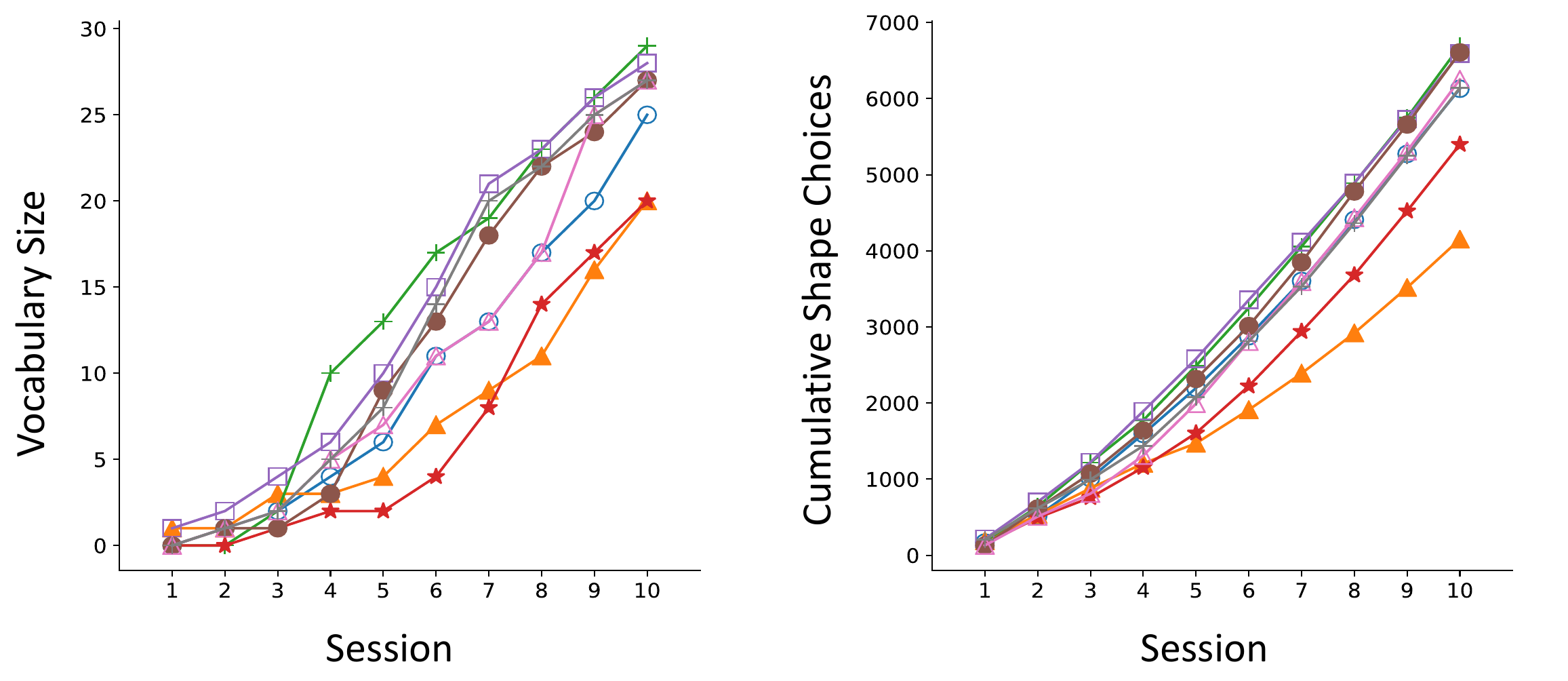}
            \end{center}
            \vspace*{-0.9em}
            \caption{CNN}
            \label{fig:learning_curves_network}
        \end{subfigure}
    \end{center}
    \vspace*{-0.7em}
    \caption{Learning curves for shape bias and vocabulary.
    (a) shows the learning curves of the 8 children participants from \cite{GershkoffStowe2004}. Participants were studied over the course of 5-8 lab sessions. Curves are shown for vocabulary size (left) and cumulative shape choices (right). Here, vocabulary includes all noun types.
    (b) shows analogous plots for our CNNs. 8 networks are shown, randomly sampled from the total 20 for the sake of visibility. Here, vocabulary is measured only for shape-based object names.
    }
    \label{fig:learning_curves}
    \vspace*{-1.0em}
\end{figure}

We inspect the ``early'' word learning period for our networks, defined as the period in which the average vocabulary size across the 20 networks is less than or equal to $2/3$ the total number of count nouns.
Beyond this period, which we find to include the first 30 training epochs, the network's learning begins to flatten. We divide this period into 10 ``sessions,'' evenly spaced by 3 epochs. The learning curves of our networks are shown in Fig. \ref{fig:learning_curves_network}.
We compute correlation metrics for our networks that are analogous to those of the child study.
Looking at increases across the sessions of a single network (metric {\bf 1}), we find an average correlation of $r=.53$ between increase in cumulative shape choices and increase in cumulative count nouns [$p<.05$ for each]. Further, looking at average increases across the entire 10-session period for each network (metric {\bf 2}), we find a correlation of $r=0.76$ [$p< .001$] across the 20 networks.

These analyses confirm that the dynamics of shape bias acquisition and early word learning show a considerable dependency on one another in our CNNs, a phenomenon that is mirrored in the early word learning of human children.

\vspace*{-0.5em}
\section{Conclusion}
\vspace*{-0.3em}
Using a set of controlled synthetic experiments, our work provides novel insights about the environmental conditions that enable learning-to-learn in neural networks. Building on the work of \cite{Colunga2005} and \cite{Ritter2017}, Experiment 1 showed that simple neural networks can learn a shape bias from stimuli presented as abstract bit patterns with as few as 3 examples of 4 categories. Experiment 2 showed that simple convolutional neural net architectures trained on high-dimensional images can learn a shape bias with as few as 6 examples of 8 object categories. Although Hierarchical Bayesian Models (HBMs) are often noted for their data efficiency, our results indicate that neural networks can approach both HBMs \citep{Kemp2007} and children \citep{Smith2002} in the amount of data required to develop a shape bias. Moreover, we show that the complexity of the data (e.g., binary patterns vs. synthetic images) influences the dynamics of learning, and that neural networks are a powerful tool for understanding these types of interactions.

The development of the shape bias in children is known to correlate with accelerated word learning \citep{GershkoffStowe2004}, a phenomenon that Experiment 3 confirmed can be mirrored in neural networks. One implication of this finding is that it may be possible to train large-scale image recognition models more efficiently after initializing these models with shape bias training. In future work, we hope to investigate this hypothesis with ImageNet-scale DNNs, using an initialization framework designed with the intuitions garnered here.

\nocite{Dubuisson1994}

\bibliographystyle{apacite}

\setlength{\bibleftmargin}{.125in}
\setlength{\bibindent}{-\bibleftmargin}

\bibliography{citations}

\newpage
\section{SUPPLEMENTAL MATERIAL}
\subsection{SM 1: Selection of network architectures and training parameters}
\label{sec:sm1}
\subsubsection{Architecture.} The MLP architecture for Experiment 1 was chosen ad-hoc before running any experiments. The network receives a 60-dimensional input, and thus, we chose a hidden layer size of 30 units to reduce this dimensionality by a factor of 2. L2 regularization was critical to the performance of the network when small training sets were provided. For Experiment 2, we chose the minimal CNN architecture that could effectively learn the image classification task that it was assigned. We generated a large dataset of 30 categories and 20+ examples per category. Then, we started with a large CNN and iteratively reduced the number of parameters until the minimal architecture was found. L2 regularization was again critical to model performance.

\subsubsection{Training parameters.} For both the MLP and the CNN, we train the network to minimize negative log-likelihood loss, using stochastic gradient descent (SGD) with the RMSprop update rule and a typical batch size of 32. There are a few exceptions to this batch size: when the training set is very small, we adjust the batch size to ensure there are at least 5 training batches. Thus, for a training set with $N$ categories and $K$ examples per category (a total of $N*K$ training points), we use a batch size of min(32, $\frac{N*K}{5}$). The number of training epochs was chosen such that the network loss reaches an asymptote for each the MLP and CNN. Training loss is monitored and used to save the best model.

\subsection{SM 2: Experiment 2 perceptual sensitivity tests}
\label{sec:sm2}
As in Experiment 1, for Experiment 2 we parametrically manipulate the stimuli to analyze the network's sensitivity to changes along different stimulus dimensions, using a CNN trained with $N$=30 \& $K$=10. Distance in shape space is quantified as the Modified Hausdorff Distance \citep{Dubuisson1994} between the shape pair. In color space, physical distance is quantified using the cosine similarity of the RGB vector pair. Beginning with an exemplar object stimuli, we sample 50 secondary shapes and order them by their distance from the exemplar. We then modify the shape of the exemplar parametrically by stepping along this list, recording network similarities between the original and modified versions in each case. A mirroring experiment is then performed with color; in each case, only 1 attribute is altered at a time. Results are shown in Fig. 4. As with the MLP, our CNN's selection preferences show a clear parametric dependency on shape, and a much weaker dependency on color

\begin{figure}[t!]
    \begin{center}
    \begin{subfigure}[b]{0.18\textwidth}
        \begin{center}
            \includegraphics[width=\textwidth]
            {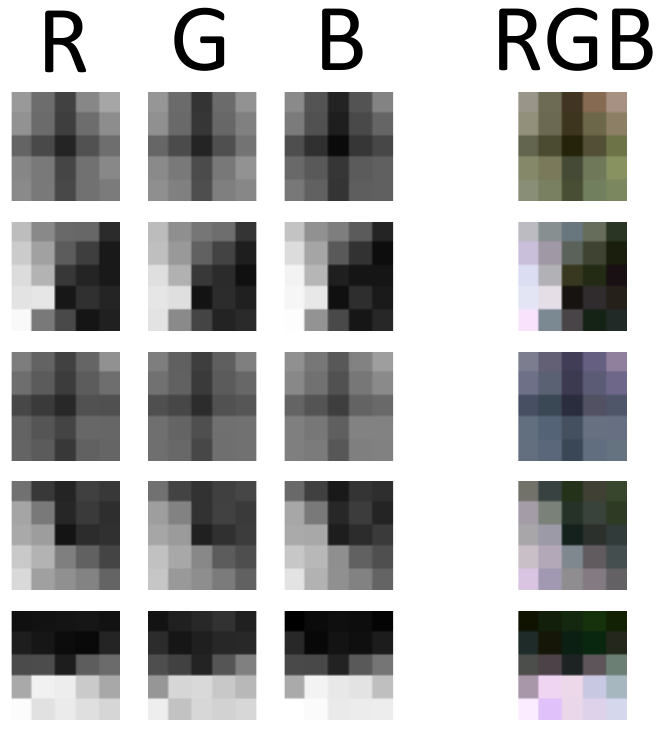}
            \caption{Shape}
            \label{fig:filters_shape}
        \end{center}
    \end{subfigure} \hspace{0.05\textwidth}
    \begin{subfigure}[b]{0.18\textwidth}
        \begin{center}
            \includegraphics[width=\textwidth]
            {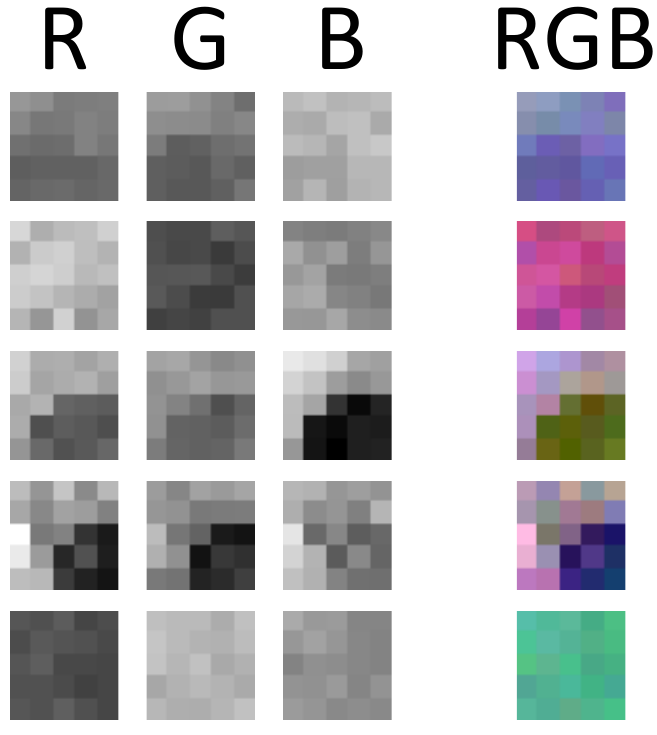}
            \caption{Color}
            \label{fig:filters_color}
        \end{center}
    \end{subfigure}
    \caption{Visualizing RGB channels of learned first-layer convolution filters. (a) shows the filters of our CNN trained with explicit shape bias training ($N$=50 \& $K$=18). Each row corresponds to 1 of the 5 filters. The first 3 channels are shown in the 'R', 'G' and 'B' columns, respectively.  These 3 channels are shown together in a 4th column, labeled 'RGB'. (b) shows mirroring filters for our CNN trained to label objects with category names based on color. In both (a) and (b), only channels 1-3 of the 4 are shown.}
    \vspace*{-1.5em}
    \label{fig:filter_visualizations}
    \end{center}
\end{figure}

\begin{figure*}[t!]
    \begin{center}
        \begin{subfigure}[b]{0.48\textwidth}
            \begin{center}
                \includegraphics[width=0.98\textwidth]
                {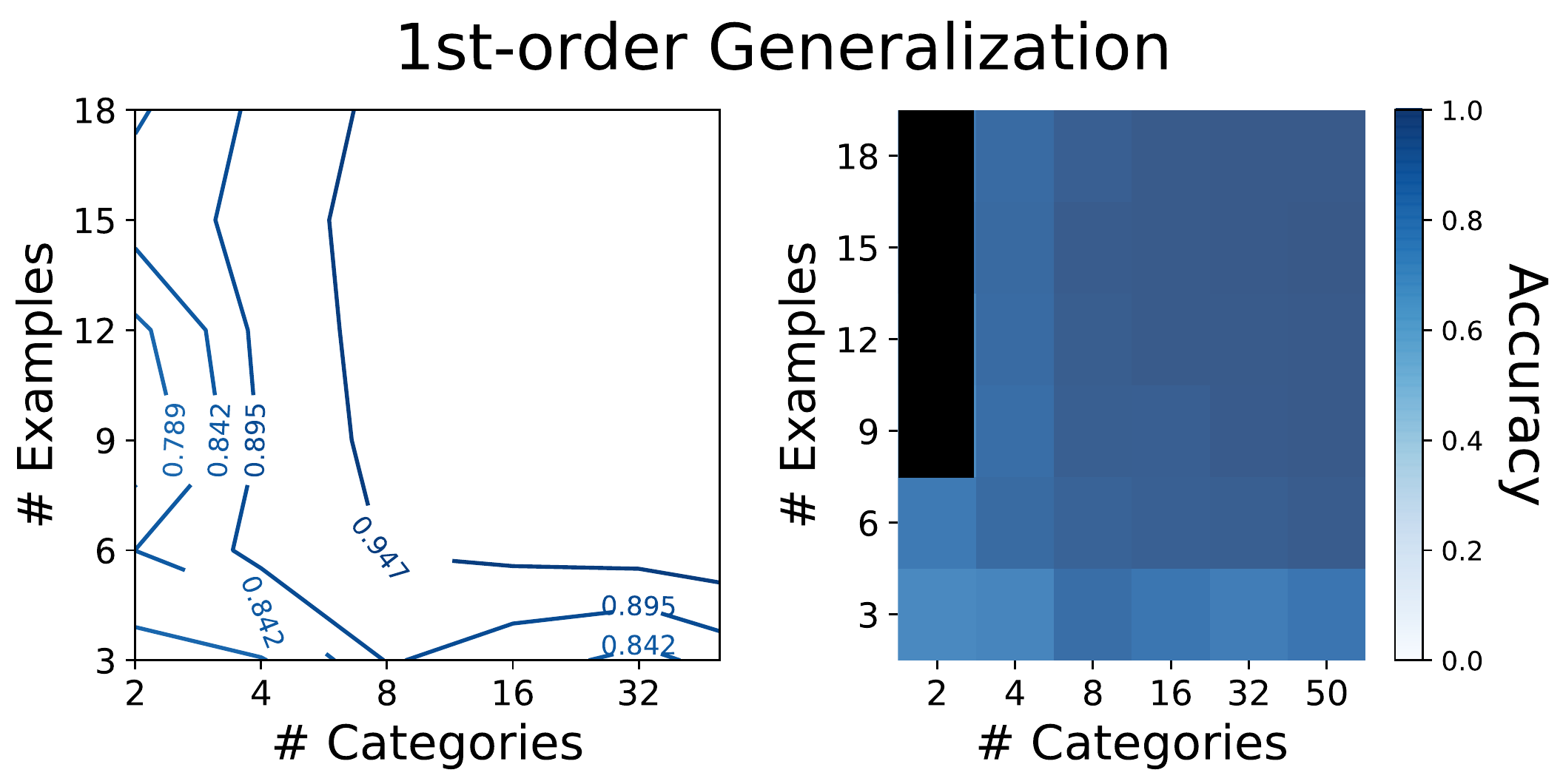}
            \end{center}
        \end{subfigure}
        \begin{subfigure}[b]{0.48\textwidth}
            \begin{center}
                \includegraphics[width=0.98\textwidth]
                {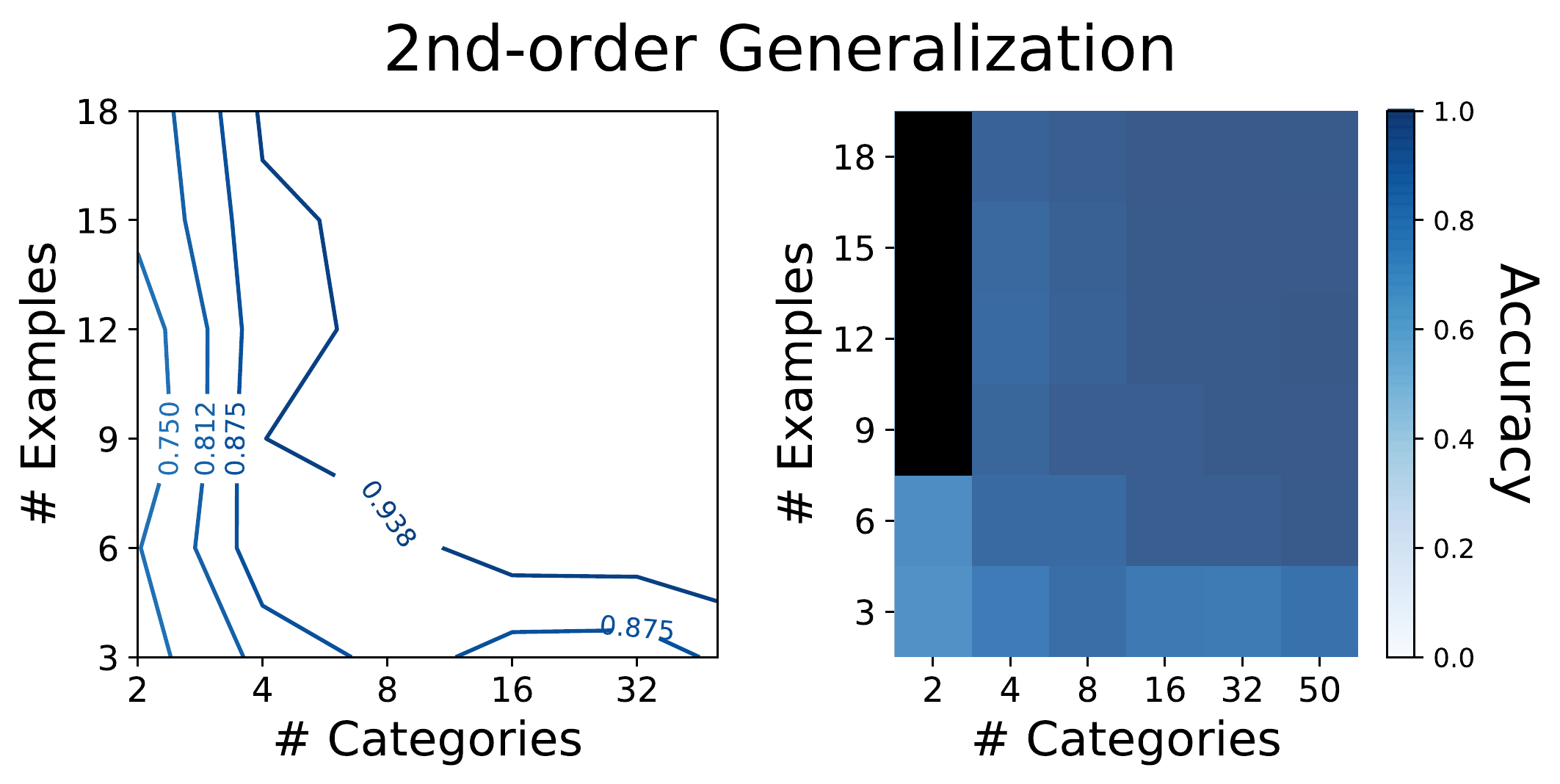}
            \end{center}
        \end{subfigure}
    \end{center}
    \caption{CNN generalization results for color bias training. The network is trained to label objects with category names based on color. In this case, the generalization tests evaluate the fraction of times that the color match is selected. Results in each grid show the average of 10 training runs.}
    \label{fig:cnn_gen_results_color}
\end{figure*}

\subsection{SM 3: Experiment 2 color bias training}
\label{sec:sm3}
For the sake of comparison, in Experiment 2 we also trained our CNN to label objects with names organized by color.
Our goal was to compare the required sample complexity for color bias training with that of shape bias training, and to evaluate whether color bias development follows a similar 2-step process.
All dataset parameters mirrored those of shape training, except that the object labels were aligned with the color attribute of each training image.
Performance on the generalization tests was measured as the fraction of trials for which the network selects the color match.
Results for CNN color bias training are shown in Fig. \ref{fig:cnn_gen_results_color}.
Notably, the color-trained CNN requires a smaller sample complexity to achieve 0.7 accuracy on the \nth{2}-order test, reaching a score of 0.73 with $N$=2 \& $K$=3.
Furthermore, this network does not appear to follow the 2-step process of bias development; results for \nth{1}- and \nth{2}-order generalizations look near-identical to one another.
In order to identify a stimulus as a member of a particular color category, the network needs only to find a single pixel of that color, a task that is much simpler than representing and identifying shape.
Representing color requires a simple 3D space.
By learning to isolate and preserve this space in the hidden layers, the network can easily generalize to novel colors, hence the early \nth{2}-order results.
We inspected the learned representations of both a shape-trained and a color-trained CNN, trained with $N$=50 \& $K$=18, by visualizing the first-layer convolution filters of each network (Fig. \ref{fig:filter_visualizations}).
As we would expect, filters of the shape-trained CNN look identical across R, G and B channels, as this network needs no sensitivity to color. In contrast, filters of the color-trained CNN vary across channels, indicating that the network has learned a selectivity for color.

\subsection{SM 4: Experiment 3 network details}
\label{sec:sm4}
\begin{figure}[t!]
    \begin{center}
        \includegraphics[width=0.5\textwidth]{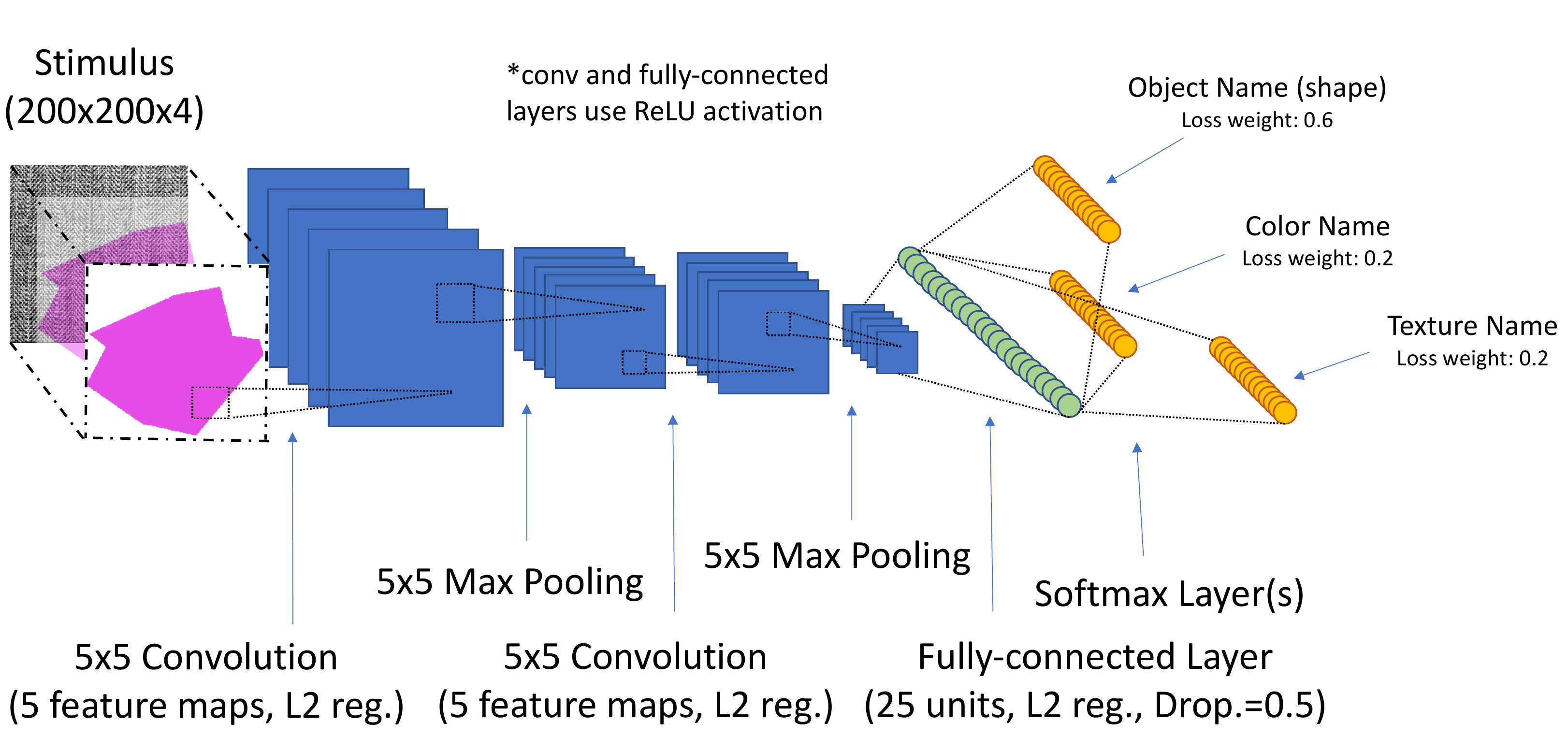}
    \end{center}
    \caption{CNN architecture for Experiment 3. The architecture mimics the original CNN of Experiment 2, with the exception of the softmax layer. Here, there are 3 softmax layers (1 for each shape, color and texture), each of which extends from the fully-connected layer.}
    \label{fig:cnn_experiment3}
\end{figure}

In Experiment 3, we use a slightly modified version of the CNN from Experiment 2 (see Fig. \ref{fig:cnn_experiment3}). We train our CNN to simultaneously label the object's name, which correlates with shape, as well as its color and texture names. The CNN thus has 3 softmax layers, each of which extends from the same fully-connected layer, and each of which has its own negative log-likelihood loss function. The training loss is computed as a weighted average of the 3 losses, with weights of 0.6, 0.2 and 0.2 assigned to shape, color and texture, respectively:
\begin{equation*}
Loss = \\
0.6*shape\_loss +
0.2*color\_loss +
0.2*texture\_loss
\end{equation*}

\end{document}